\documentclass[11pt,a4paper]{article}
\usepackage[hyperref]{emnlp2018}
\usepackage{times}
\usepackage{latexsym}
\usepackage[norelsize]{algorithm2e}
\usepackage{url}

\usepackage{extarrows}
\usepackage{amssymb,amsfonts}
\usepackage{latexsym}
\usepackage{multirow}
\usepackage{arydshln}
\usepackage{amsmath}
\usepackage{rotating}
\usepackage{color}
\usepackage{subfigure}
\usepackage{enumitem}
\aclfinalcopy % Uncomment this line for the final submission
\newcommand{\ignore}[1]{}

\title{\modelname: A Two-Wing Optimization Strategy \\for Evidential Claim Verification}

\author{Wenpeng Yin, Dan Roth \\
  University of Pennsylvania\\
  {\tt \{wenpeng,danroth\}@seas.upenn.edu}}

\date{}

\newcommand{\dataname}{\textsc{FEVER}}
\newcommand{\modelname}{\textsc{TwoWingOS}}

\begin{document}
\maketitle
\begin{abstract}
Determining whether a given claim is supported by evidence is a fundamental NLP problem that is best modeled as Textual Entailment. However, given a large collection of text, finding evidence that could support or refute a given claim is a challenge in itself, amplified by the fact that different evidence might be needed to support or refute a claim. Nevertheless, most prior work decouples evidence identification from determining the truth value of the claim given the evidence.  

We propose to consider these two aspects jointly. We develop  \modelname\enspace (\textbf{two}-\textbf{wing}  \textbf{o}ptimization \textbf{s}trategy), a system that, while identifying appropriate evidence for a claim, also determines whether or not the claim is supported by the evidence.
Given the claim, \modelname\enspace attempts to identify a subset of the evidence candidates; given the predicted evidence, it then attempts to determine the truth value of the corresponding claim. We treat this challenge as coupled optimization problems, training a joint model for it.
\modelname\enspace offers two advantages: (i) Unlike pipeline systems, it facilitates flexible-size evidence set, and (ii) Joint training  improves both the claim verification and the evidence identification. Experiments on a  benchmark dataset show state-of-the-art performance.\footnote{\url{cogcomp.org/page/publication_view/847}}
\end{abstract}

\section{Introduction}
\begin{figure}
\centering
\includegraphics[width=0.48\textwidth]{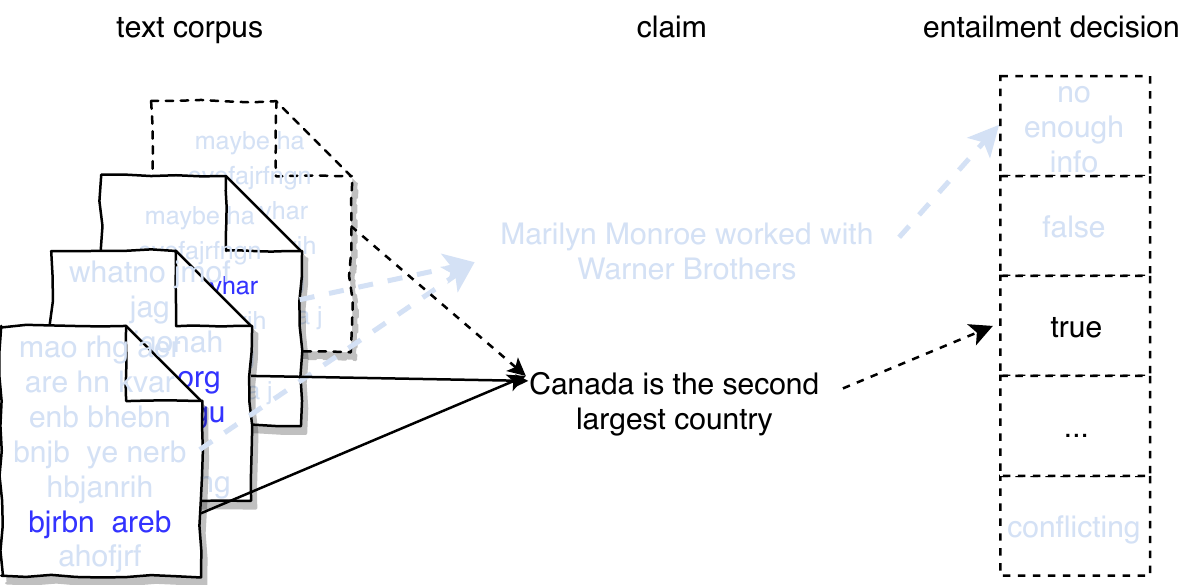}
\caption{Illustration of evidential claim verification task. For a claim, we determine its truth value by evidence identified from a text corpus.}\label{fig:taskfig}
\end{figure}

A claim, e.g., ``Marilyn Monroe worked with Warner Brothers'',  is an assertive sentence that may be true or false. While the task of claim verification {\em will not} tell us the absolute truth of this claim, it is expected to determine whether the claim is supported by evidence in a given text collection. Specifically, given a claim and a text corpus, {\em evidential claim verification}, demonstrated in Figure \ref{fig:taskfig}, aims at identifying text snippets in the corpus that  act as evidence that supports or refutes the claim. 

This problem has broad applications. For example,  knowledge bases (KB), such as Freebase \cite{bollacker2008freebase}, YAGO \cite{DBLPSuchanekKW07}, can be augmented with a new relational statement such as ``(Afghanistan, is\_source\_of, Kushan Dynasty)''. This needs to be first verified by a claim verification process  and supported by evidence \cite{DBLPRothSV09,DBLPChagantyPLM17}. More broadly,  claim verification is a key component in any technical solution addressing recent concerns about the trustworthiness of online content \cite{DBLPVydiswaranZR11,DBLPPasternackR13,DBLPHovyBVH13}.
%has exacerbated long-standing concerns,  sharpening desires for technical solutions to assess the reliability of claims in a wild media environment.
% More broadly,  The furor over  misleading content    online has exacerbated long-standing concerns about knowledge acquisition in a wild media environment. Usually, we need to assess the truthfulness of claims made by public figures such as investors, politicians, etc., and even newspaper headlines from some unknown media, by consulting a variety of sources. This is time-consuming,   sharpening desires for technological solutions.  
In both scenarios, we care about whether or not a claim holds, and seek reliable evidence in support of this decision.

Evidential claim verification requires that we address three challenges.
First,  to locate text snippets in the given corpus that can  potentially be  used to determine the truth value of the given claim.  
% This differs from sentence-to-sentence similarity because it is not the case that a highly similar sentence is more likely to support or refute a claim.
This  differs from the conventional textual entailment (TE) problem~\cite{DRSZ13} as here we first look for the premises given a hypothesis. Clearly, the evidence one seeks depends on the claim, as well as on the eventual entailment decision -- the same claim would require different supporting than refuting evidence. This motivates us to develop an approach that can transfer knowledge from claim verification to  evidence identification. Second, the evidence for a claim might require aggregating information from multiple sentences and even multiple documents (rf. \#3 in Table \ref{tab:errorexample}). Therefore, a {\em set}, rather than a collection of independent text snippets, should be chosen to act as evidence.
And, finally, in difference from TE, given a set of evidence sentences as a premise, the truth value of the claim should depend on {\em all of the evidence}, rather than on a single sentence there. 
\begin{figure}
\centering
\includegraphics[width=0.47\textwidth]{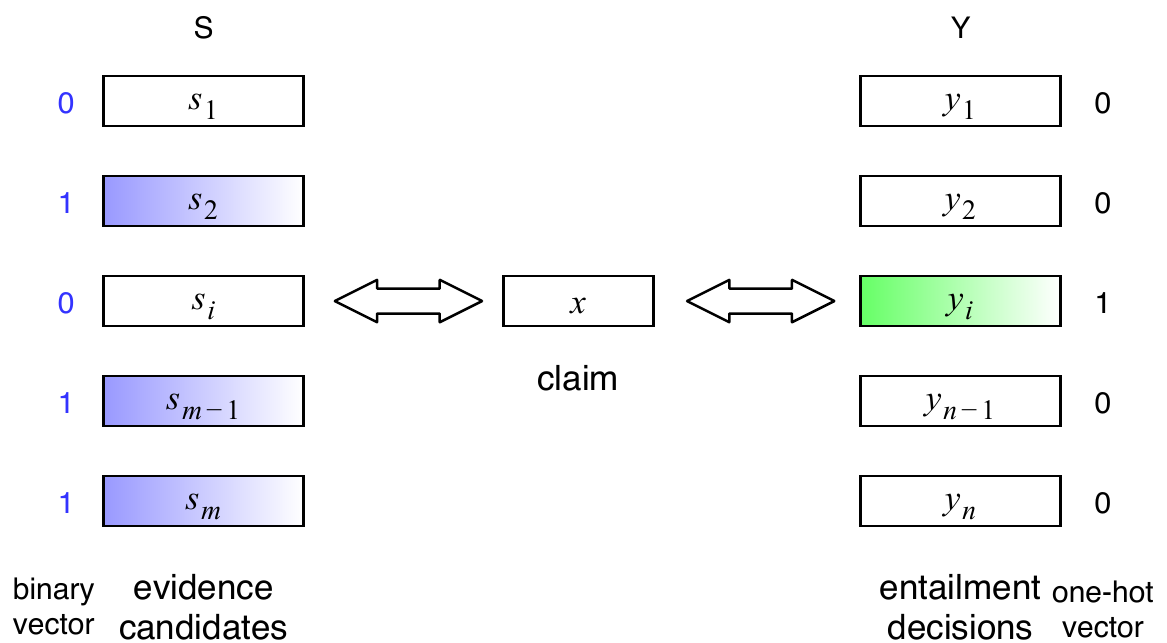}
\caption{\modelname, a generic two-wing optimization framework. A subset of the evidence candidates $S_e$ = \{$s_1,\ldots, s_{m-1}, s_m$\} is chosen via a binary vector (left), and an n-valued  entailment decision $y_i\in Y$ is chosen (right), with respect to the claim $x$.}\label{fig:bigp}
\end{figure}

The discussion above suggests that claim verification and evidence identification are tightly coupled. Claim should influence the identification of appropriate evidence, and ``trusted evidence boosts the claim's veracity'' \cite{DBLPVydiswaranZR11}.
%and claims under the training for entailment know more about which kind of evidence they are looking for. This motivates us to tackle them in a joint learning process. 
Consequently, we propose \modelname, a two-wing optimization strategy\footnote{By ``two-wing optimization'', we mean that the same object, i.e., the claim, is mapped
%learned 
into two target spaces in a joint optimization scheme.}, to support this process. As shown in Figure \ref{fig:bigp}, we consider a set of sentences $S$ as the candidate evidence space, a claim $x$, and a decision space $Y$ for the claim verification. In the optimal condition, a one-hot vector over $Y$ indicates which decision to make towards the claim, and a binary vector over $S$ indicates a subset of sentences $S_e$ (in blue in Figure \ref{fig:bigp}) to act as evidence. 

Prior work mostly approached this problem as a pipeline procedure -- first, given a claim $x$, determine $S_e$  by some similarity matching; then, conduct textual entailment over ($S_e$, $x$) pairs. Our framework, \modelname, optimizes the two subtasks jointly, so that both claim verification and evidence identification can enhance each other. \modelname\enspace is a generic framework making use of a shared representation of the claim to co-train evidence identification and claim verification.

\modelname\enspace is tested on the FEVER benchmark \cite{DBLPfever05355}, showing $\approx$30\% $F_1$ improvement for evidence identification, and $\approx$23\% accuracy  increase in claim verification. Our analysis shows that (i) entity mentions in claims provide a strong clue for retrieving relevant passages; (ii) composition of evidence clues across sentences  helps claim verification; and that (iii) the joint training scheme provides significant benefits of a pipeline architecture. 
\ignore{
In this work, we contribute in twofold:
\begin{itemize}
[leftmargin=*]
\setlength\itemsep{0.01em}
\item We achieve an outperforming wiki-page retrieval component for claims, and release the retrieval results;
\item To our knowledge, the two-wing optimization structure in this topic is rarely handled in a joint way before. We propose \modelname\enspace for jointly reasoning the two coupled subtasks. After document retrieval, the whole system can be trained end-to-end.
% \item \textcolor{red}{something else}
\end{itemize}
}

\section{Related Work}

% \subsection{Fact-checking}
Most work focuses on the dataset construction  
% , such as \cite{DBLPVlachosR14,DBLPFerreiraV16,DBLPWang17,WinNT}, 
while lacking advanced models to  handle the problem. \newcite{DBLPVlachosR14} propose and define the ``fact checking'' problem, without a concrete solution. \newcite{DBLPFerreiraV16} release the dataset ``Emergent'' for rumor debunking. Each claim is accompanied by an article headline as evidence. Then a three-way logistic regression model is used over some rule-based features. No need to search for evidence. \newcite{DBLPWang17} release a larger dataset for fake news detection, and propose a hybrid  neural network to integrate the statement and the speaker's meta data to do classification. However, the presentation of evidences is ignored. \newcite{DBLPKobayashiIHMM17} release a similar dataset to \cite{DBLPfever05355}, but they do not consider the evaluation of evidence reasoning.

Some work mainly pays  attention to determining whether the claim is true or false, assuming evidence facts are provided or  neglecting presenting evidence totally, e.g., \cite{DBLPngeliM14} -- given a database of true facts as premises,  predicting whether an unseen fact is true and should belong to the database by natural logic inference. Open-domain question answering (QA) against a  text corpus \cite{yinHCQA,DBLPChenFWB17,DBLPWangYGWKZCTZJ18} can also be treated as claim verification problem, if we treat (question, correct answer) as a claim. However, little work has studied how well a QA system can identify all the answer evidence.

Only a few works considered  improving the evidence presentation in claim verification problems. \newcite{DBLPRothSV09} introduce the task of Entailed Relation Recognition -- given a set of short paragraphs and a relational fact in the triple form of (argument$_1$, relation, argument$_2$),  finding the paragraphs that can entail this fact. They first use Expanded Lexical Retrieval to  rank and keep the top-$k$ paragraphs as candidates, then build a TE classifier over each (candidate, statement) pair.  The work directly related  to us is by \newcite{DBLPfever05355}. Given claims and a set of Wikipages, \newcite{DBLPfever05355} use a retrieval model based on TF-IDF  to locate top-5 sentences in top-5 pages   as evidence, then utilize a neural entailment model to classify (evidence, claim) pairs. 

In contrast, our work tries to optimize the claim verification as well as the evidence identification in a joint training scheme, which is more than just supporting or refuting the claims.

% \subsection{QA-driven fact verification}

% \newcite{scitail} and \newcite{wenpengacl2018} study a textual entailment task derived from a multi-choice question answering problem. Each hypothesis there is constructed by converting a (question, correct answer) into a statement. Compared to the work here, we require to infer the evidence sentences rather than infer merely the entailment labels.

% \newcite{DBLPChenFWB17} study open-domain question answering against Wikipedia. First a document retrieval module is developed to control the search space. Then a deep neural network is used to detect the answer span for a question. However, the answer (i.e., evidence) comes from a single piece of text.

% \newcite{DBLP05116} study open-domain question answering where a question may get answered by evidences from multiple passages. If we treat (question, answer candidate) as a statement, then this QA problem can also act as fact checking task.

% \textcolor{red}{put in different position?} In data mining community, \newcite{DBLPVydiswaranZR11}  model the trustworthiness of claims, evidence (web pages) and news cites in a three-tier graph and explore a trust propagation framework. The nodes in the graph are connected by some shallow feature based similarity scores. These system might work well for news claims, but tend to be fragile towards claims made by humans. Since their node connections are based on shallow features, they are less sensitive to some negations, substitutions etc. Other similar work includes \cite{DBLPYinHY07,vydiswaran2011gauging} etc.

\section{The \modelname\enspace Model}
% We use  bold uppercase, e.g.,
% $\mathbf{W}$, for matrices;
% bold lowercase,
% e.g.,
% $\mathbf{h}$,  for vectors; and non-bold lowercase for
% scalars.

Figure \ref{fig:bigp} illustrates the  two-wing optimization problem addressed in this work: given a collection of evidence candidates $S$=\{$s_1$, $s_2$, $\cdots$, $s_i$, $\cdots$, $s_m$\},  a claim $x$ and a decision set $Y$ = \{$y_1$ $\cdots$, $y_n$\}, the model \modelname\enspace predicts a binary vector $p$ over $S$ and a one-hot vector $o$ over $Y$ against the ground truth, a binary vector $q$ and a one-hot vector $z$, respectively. A binary vector over $S$ means a subset of sentences ($S_e$) act as evidence, and the one-hot vector indicates a single decision ($y_i$) to be made towards the claim $x$ given the evidence $S_e$. Next, we use two separate subsections to elaborate the process of evidence identification  (i.e., optimize $p$ to $q$) and the claim verification (i.e., optimize $o$ to $z$).

\subsection{Evidence identification}
A simple approach to identifying evidence is to detect the top-$k$ sentences that are lexically similar to the claim, as some pipeline systems \cite{DBLPRothSV09,DBLPfever05355} do. However, a claim-unaware fixed $k$  is less optimal, adding noise or missing key supporting factors, consequently limiting the performance.
% A simple approach to locate some sentences as evidence  is to set a step in the pipeline system that searches for some sentences which are lexically similar to the claim  \cite{DBLPRothSV09,DBLPfever05355}. However, it is hard to set a claim-specific threshold to select or discard a sentence. So, usually a set of top-$n$ candidates are kept, in which $n$ is the same all over the claims. This is apparently less optimal. A fixed number can not locate the evidence precisely as it will  bring noise or miss key supporting factors. Consequently, an upper bound exists for the performance of  evidence identification as well as the claim verification. 
% (iii) The same claim with a different decision $y_i\in Y$ requires different evidence. (iv) A claim may requires multiple sentences as a whole as a evidence So, we need to consider the relationship between (sentence, claim) as well between sentence candidates.

In this work, we approach the evidence by modeling sentences $S$=\{$s_1$,  $\cdots$, $s_i$, $\cdots$, $s_m$\} with  the claim $x$ as context in a supervised learning scheme. For each $s_i$, the problem turns out to be learning a probability: how likely  $s_i$ can entail the claim conditioned on other candidates as context, as shown by the blue items  in Figure \ref{fig:bigp}. 

% \begin{figure}
% %  \setlength{\belowcaptionskip}{-12pt}
% %  \setlength{\abovecaptionskip}{5pt}
% \centering
% \includegraphics[width=0.48\textwidth]{reasonevidence}
% \caption{How likely $s_i$ entails $x$ with  other sentence candidates as context.}\label{fig:reasonevi}
% \end{figure}

To start, a piece of text $t$  ($t \in S\cup\{x\}$) is represented as a sequence of
$l$ hidden states,  forming a feature map
$\mathbf{T}\in\mathbb{R}^{d\times l}$, where $d$ is the
dimensionality of hidden states. We first stack a vanilla CNN (convolution \& max-pooling) \cite{lecun1998gradient} over $\mathbf{T}$ to get a representation for $t$. As a result, each evidence candidate $s_i$ has a representation $\mathbf{s}_i$, and the claim $x$ has a representation $\mathbf{x}$. To get a probability for each $s_i$, we need first to build its \emph{claim-aware representation} $\mathbf{r}_i$.

\paragraph{Coarse-grained representation.}  We directly concatenate the representation of $s_i$ and $x$, generated by the vanilla CNN, as:
\begin{equation}
\mathbf{r}_i = [\mathbf{s}_i, \mathbf{x}, \mathbf{s}_i\cdot\mathbf{x}^T]
\end{equation}
This coarse-grained approach makes use of merely the sentence-level representations while neglecting more fine-grained interactions between the sentences and the claim. 

\paragraph{Fine-grained representation.} Instead of directly employing the sentence-level representations, here we explore claim-aware representations for each word in sentence $s_i$, then compose them as the sentence representation $\mathbf{r}_i$, inspired by the Attentive Convolution \cite{DBLP00519}. 

For each word $s^j_i$ in $s_i$, we first calculate its matching score towards each word $x^z$ in $x$, by dot product over their hidden states. Then the representation of the claim, as the context for the word $s^j_i$, is formed as:
\begin{equation}
\label{eq:extracontext}
\mathbf{c}^j_i = \sum_z \mathrm{softmax}(\mathbf{s}^j_i\cdot(\mathbf{x}^z)^T)\cdot \mathbf{x}^z
\end{equation}

Now, word $s^j_i$ has  left
context $\mathbf{s}^{j-1}_i$, right context
$\mathbf{s}^{j+1}_i$ in $s_i$, and the claim-aware context
$\mathbf{c}^j_i$ from $x$. A convolution encoder  generates its claim-aware  representation $\mathbf{i}^j_i$:
\begin{equation}
\mathbf{i}^j_i=\mathrm{tanh}(\mathbf{W}\cdot [\mathbf{s}^{j-1}_i,\mathbf{s}^j_i,\mathbf{s}^{j+1}_i,\mathbf{c}^j_i]+\mathbf{b}) 
\end{equation}
where parameters $\mathbf{W}\in\mathbb{R}^{d\times 4d}$, $\mathbf{b}\in\mathbb{R}^d$.

To compose those claim-aware word representations as the representation for sentence $s_i$, we use a max-pooling over \{$\mathbf{i}^j_i$\} along with $j$, generating $\mathbf{i}_i$. 

We use term $f_{\mathrm{int}}(s_i,x)$ to denote this whole process, so that:
\begin{equation}
\label{eq:attconv}
\mathbf{i}_i = f_{\mathrm{int}}(s_i,x)
\end{equation}

At this point, the fine-grained  representation for evidence candidate $s_i$ is:
\begin{equation}
\mathbf{r}_i = [\mathbf{s}_i, \mathbf{x}, \mathbf{s}_i\cdot\mathbf{x}^T, \mathbf{i}_i]
\end{equation}

\paragraph{Loss function.} With a claim-aware representation $\mathbf{r}_i$, the sentence $s_i$ subsequently gets a probability, acting as the evidence, $\alpha_i\in(0,1)$ via a  non-linear sigmoid function:
\begin{equation}
\label{eq:prob}
\alpha_i = \mathrm{sigmoid}(\mathbf{v}\cdot \mathbf{r}_i^\mathrm{T})
\end{equation}
where parameter vector $\mathbf{v}$ has the same dimensionality as $\mathbf{r}_i$.

In the end, all evidence candidates in $S$ have a ground-truth binary vector $q$ and the predicted probability vector $\alpha$; then loss $l_{ev}$ (``ev'': \textbf{ev}idence) is implemented as a binary cross-entropy:
\begin{equation}
l_{ev} = \sum^m_{i=1}-(q_i\log(\alpha_i)+(1-q_i)\log(1-\alpha_i))
\end{equation}

As the output of this evidence identification module, we binarize the probability vector $\alpha$ by $p_i = \textbf [\alpha_i>0.5\textbf ]$  (``$\textbf [x\textbf ]$'' is 1 if $x$ is true or 0 otherwise). $p_i$ indicates $s_i$ is evidence or not. All \{$s_i$\} with $p_i=1$ act as evidence set $S_e$.

% \begin{figure}
% %  \setlength{\belowcaptionskip}{-12pt}
% %  \setlength{\abovecaptionskip}{5pt}
% \centering
% \includegraphics[width=0.48\textwidth]{reasondecision}
% \caption{Textual entailment for reasoning claim veracity. Premise: a set of sentences (in blue); Hypothesis: claim}\label{fig:reasonclaim}
% \end{figure}

\subsection{Claim verification}
\begin{figure}[!t] 
\centering 
\subfigure[Coarse-grained representations] { 
\includegraphics[width=0.85\columnwidth]{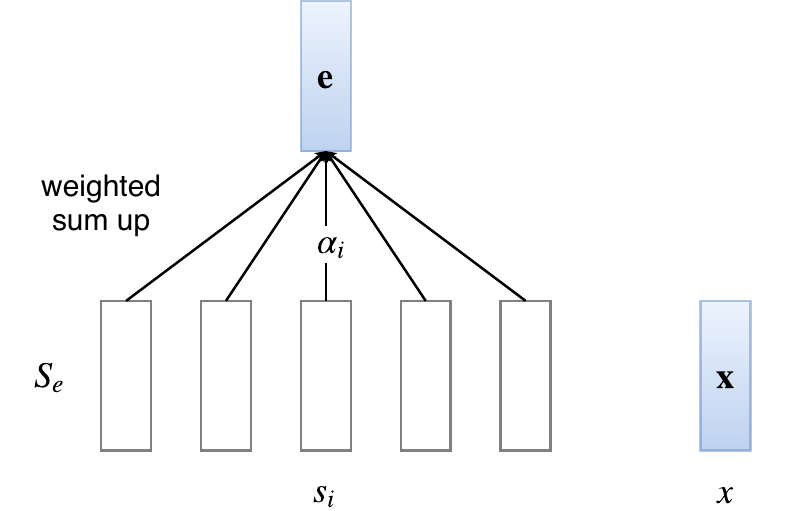} 
} 
\subfigure[Single-channel fine-grained representations] { 
\includegraphics[width=0.85\columnwidth]{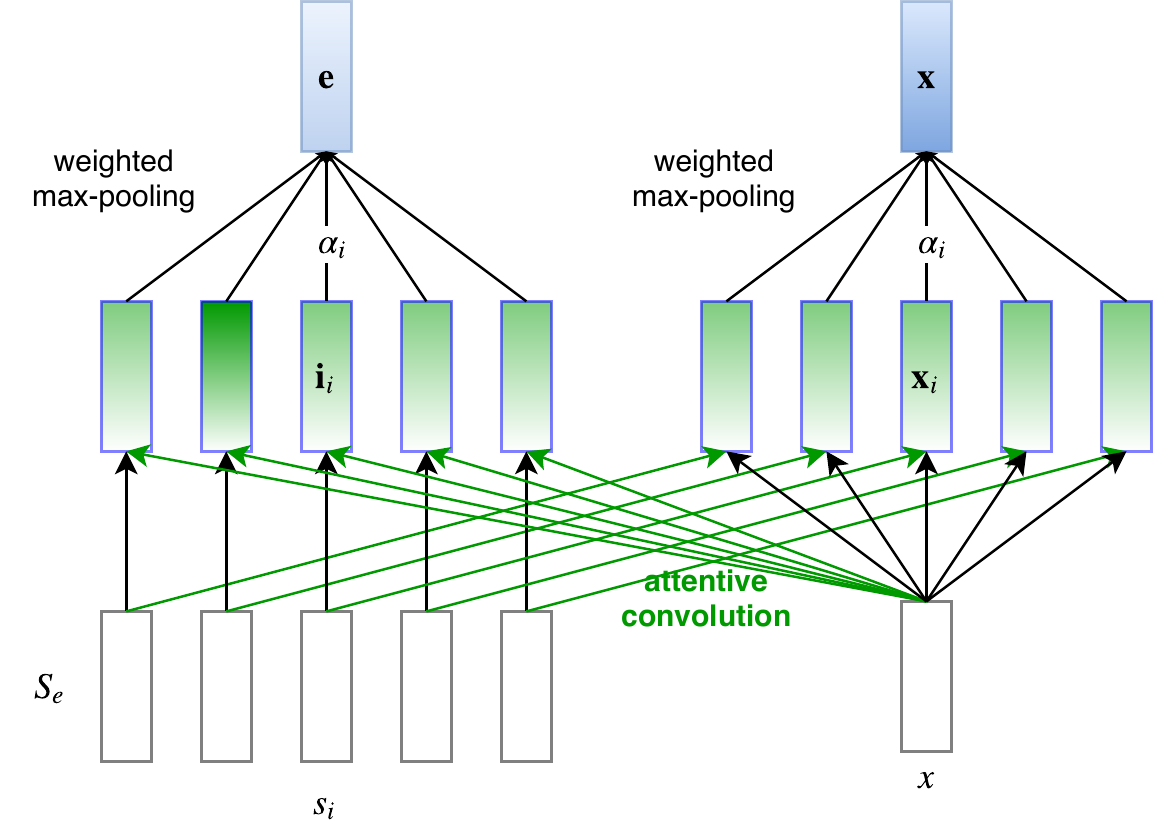} 
} 
\subfigure[Two-channel fine-grained representations] { 
\includegraphics[width=0.85\columnwidth]{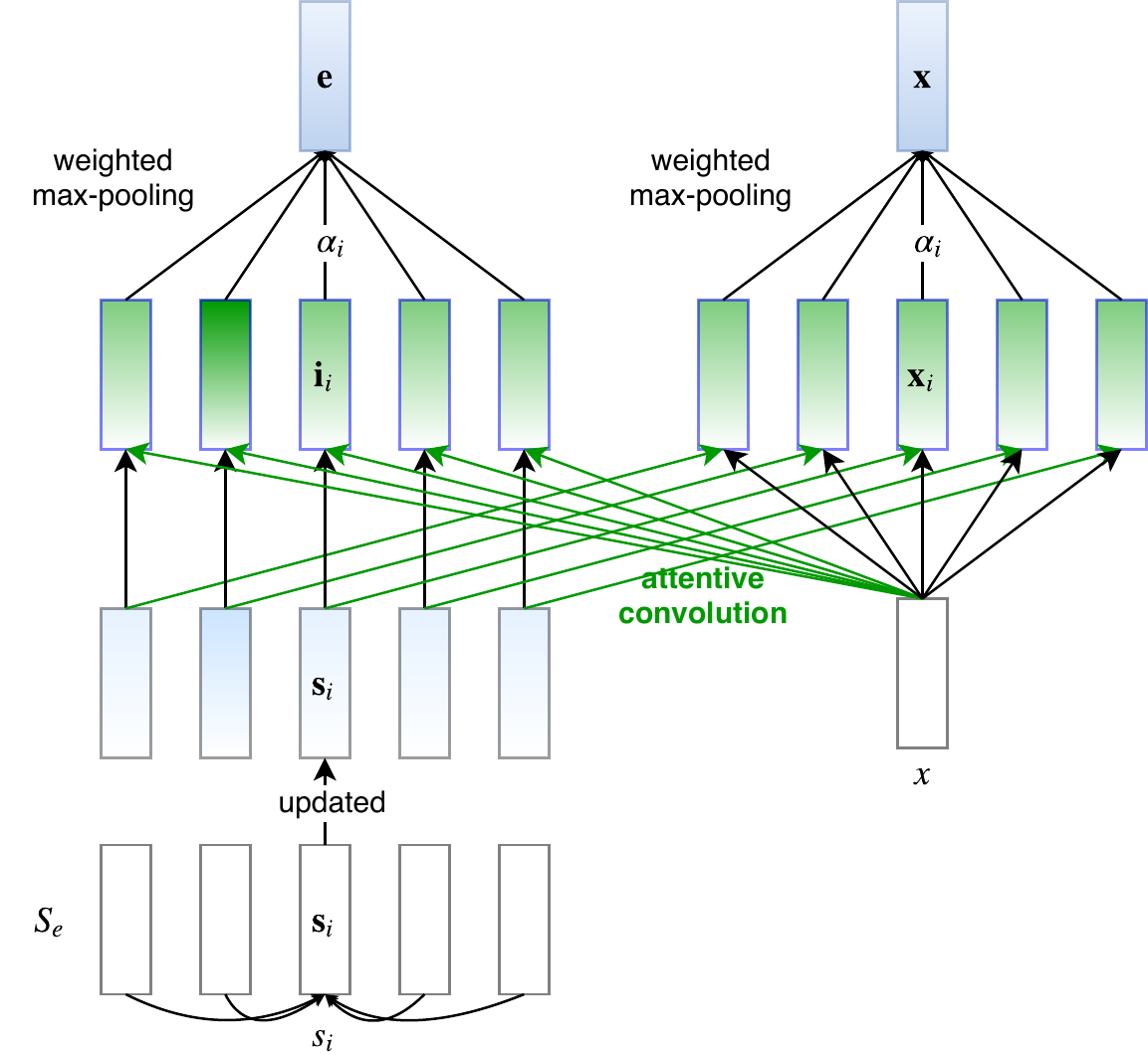} 
} 
\caption{Three representation learning methods in claim verification. Green arrows act as context in attentive convolution.} 
\label{fig:methoddetails} 
\end{figure}

As shown in Figure \ref{fig:bigp},  to figure out an entailment decision $y_i$ for the claim $x$, the evidence $S_e$ possibly consists of more than one sentence. Furthermore, those evidence sentences are not necessarily in textual order nor from the same passage. So, we need a mechanism that enables each evidence or even each word inside to be aware of the content from other evidence sentences. 
Similar to the aforementioned approach to evidence identification, we come up with three  methods, with different representation granularity, to learn a representation for  ($S_e$, $x$), i.e., the input  for claim verification, shown in Figure \ref{fig:methoddetails}.

\paragraph{Coarse-grained representation.} In this case, we treat $S_e$ as a whole, constructing its representation $\mathbf{e}$ by  summing up the representations of all sentences in $S_e$ in a weighted way:
\begin{equation}
\label{eq:co}
\mathbf{e} = \sum^m_{i=1}\alpha_i\cdot p_i\cdot\mathbf{s}_i 
\end{equation}
where $\alpha_i$, from Equation \ref{eq:prob}, is the probability of $s_i$ being the evidence.

Then the ($S_e$, $x$) pair gets a  coarse-grained concatenated representation: $[\mathbf{e}, \mathbf{x}]$. It does not model  the interactions within the evidence nor the interactions between the evidence and the claim. Based on our experience in evidence identification module, the representation  of a sentence is better learned by composing context-aware word-level representations. Next, we introduce how to learn fine-grained representation for the ($S_e$, $x$) pair.

\paragraph{Single-channel fine-grained representation.}
By ``single-channel,'' we mean  each sentence $s_i$ is aware of the claim $x$ as its single context. 

For a single pair ($s_i$, $x$), we  utilize the function $f_{\mathrm{int}}()$ in Equation \ref{eq:attconv} to build the fine-grained representations for both $s_i$ and  $x$, obtaining $\mathbf{i}_i=f_{\mathrm{int}}(s_i, x)$ for $s_i$ and $\mathbf{x}_i=f_{\mathrm{int}}(x, s_i)$ for $x$. 

For  ($S_e$, $x$), we compose all the \{$\mathbf{i}_i$\} and all the \{$\mathbf{x}_i$\} along with $i$, via a weighted max-pooling:
\begin{align}
\label{eq:single1}
\mathbf{e} &= \mathrm{maxpool}_i(\alpha_i\cdot p_i\cdot\mathbf{i}_i)\\\label{eq:single2}
\mathbf{x} &= \mathrm{maxpool}_i(\alpha_i\cdot p_i\cdot\mathbf{x}_i)
\end{align}

This weighted max-pooling ensures that the sentences with higher probabilities of being evidence have a higher chance to present their features. 
As a result,  ($S_e$, $x$) gets a concatenated representation: [$\mathbf{e}$, $\mathbf{x}$]

\paragraph{Two-channel fine-grained representation.}
By ``two-channel,'' we mean that each evidence $s_i$ is aware of two kinds of context, one from the  claim $x$, the other from the remaining evidences.

Our first step is to accumulate evidence clues within $S_e$. To start, we concatenate all sentences in $S_e$ as a fake long sentence $\hat{S}$ consisting of hidden states \{$\hat{\mathbf{s}}$\}. Similar to Equation \ref{eq:extracontext}, for each word $s^j_i$ in sentence $s_i$, we \emph{accumulate all of its related clues} ($\mathbf{c}^j_i$) from $\hat{S}$ as follows:

\begin{equation}
\label{eq:evicontext}
\mathbf{c}^j_i = \sum_z \mathrm{softmax}(\mathbf{s}^j_i\cdot(\hat{\mathbf{s}}^z)^T)\cdot \hat{\mathbf{s}}^z
\end{equation}

Then we update  $\mathbf{s}^j_i$, the representation of word $s^j_i$, by element-wise addition:
\begin{equation}
\label{eq:evicontext}
\mathbf{s}^j_i=\mathbf{s}^j_i\oplus\mathbf{c}^j_i
\end{equation}

This step enables the word $s^j_i$ to ``see''  all related clues from $S_e$. The reason we add $\mathbf{s}^j_i$ and $\mathbf{c}^j_i$ is motivated by a simple experience: Assume the claim ``Lily lives in the biggest city in Canada'', and one sentence contains a clue ``$\cdots$ Lily lives in Toronto $\cdots$'' and another sentence contains a clue ``$\cdots$ Toronto is  Canada's largest city$\cdots$''. The most simple yet effective approach to aggregating the two clues is to sum up their representation vectors \cite{DBLPBlacoeL12} (we do not concatenate them, as those clues have no consistent textual order across different $\mathbf{s}^j_i$).

After  updating the representation of each word in $s_i$, we perform the aforementioned ``single-channel fine-grained representation'' between the updated $s_i$ and the claim $x$, generating [$\mathbf{e}$, $\mathbf{x}$].

\paragraph{Loss function.} For the claim verification input ($S_e$, $x$), we forward its representation [$\mathbf{e}$, $\mathbf{x}$] to a logistic regression layer in order to infer a probability distribution $o$ over the label space $Y$: 
\begin{equation}
\mathbf{o} = \mathrm{softmax}(\mathbf{W}\cdot [\mathbf{e}, \mathbf{x}]+\mathbf{b})
\end{equation}
where $\mathbf{W}\in\mathbb{R}^{n\times 2d}$, $\mathbf{b}\in\mathbb{R}^n$

The loss $l_{\mathrm{cv}}$ (``cv'': claim verification) is implemented as negative log-likelihood:
\begin{equation}
l_{\mathrm{cv}} = -\log(\mathbf{o}\cdot\mathbf{z}^T)
\end{equation}
where $\mathbf{z}$ is the ground truth one-hot label vector for the claim $x$ on the space $Y$.

\subsection{Joint optimization}
Given the loss $l_\mathrm{ev}$ in evidence identification and the loss $l_\mathrm{cv}$ in claim verification, the overall training loss is represented by:
\begin{equation}
l = l_{\mathrm{ev}} + l_{\mathrm{cv}}
\end{equation}

To ensure that we jointly train the two coupled subtasks with intensive knowledge communication instead of simply putting two pipeline neural networks together, our \modelname\enspace has following configurations:

\textbullet\enspace  Both subsystems share the same set of word embeddings as parameters; the vanilla CNNs for learning sentence and claim representations share parameters as well.

\textbullet\enspace  The output binary vector $p$ by the evidence identification module is forwarded to the module of claim verification, as shown in Equations \ref{eq:co}-\ref{eq:single2}.

\textbullet\enspace  Though the representation of a claim's decision $y_i$ is not  put explicitly into the module of evidence identification, the claim's representation $\mathbf{x}$ will be fine-tuned by the $y_i$, so that the evidence candidates  can  get adjustment from the decision $y_i$, since the claims are shared by two modules. 

\begin{table}
% \setlength{\tabcolsep}{3pt}
%  \setlength{\belowcaptionskip}{-5pt}
%  \setlength{\abovecaptionskip}{5pt}
%  \small
  \centering
  \begin{tabular}{l|ccc}
   &  \#\textsc{Supported}  & \#\textsc{Refuted}  & \#NEI\\\hline
	train & 80,035 & 29,775 & 35,639\\
    dev & 3,333 & 3,333 & 3,333\\
    test & 3,333 & 3,333 & 3,333\\
%     total & 27,026 & 10,101 & 16,925

\end{tabular}
\caption{Statistics of claims in \dataname\enspace dataset}\label{tab:data}
\end{table}
\begin{figure}
\centering
\includegraphics[width=0.47\textwidth]{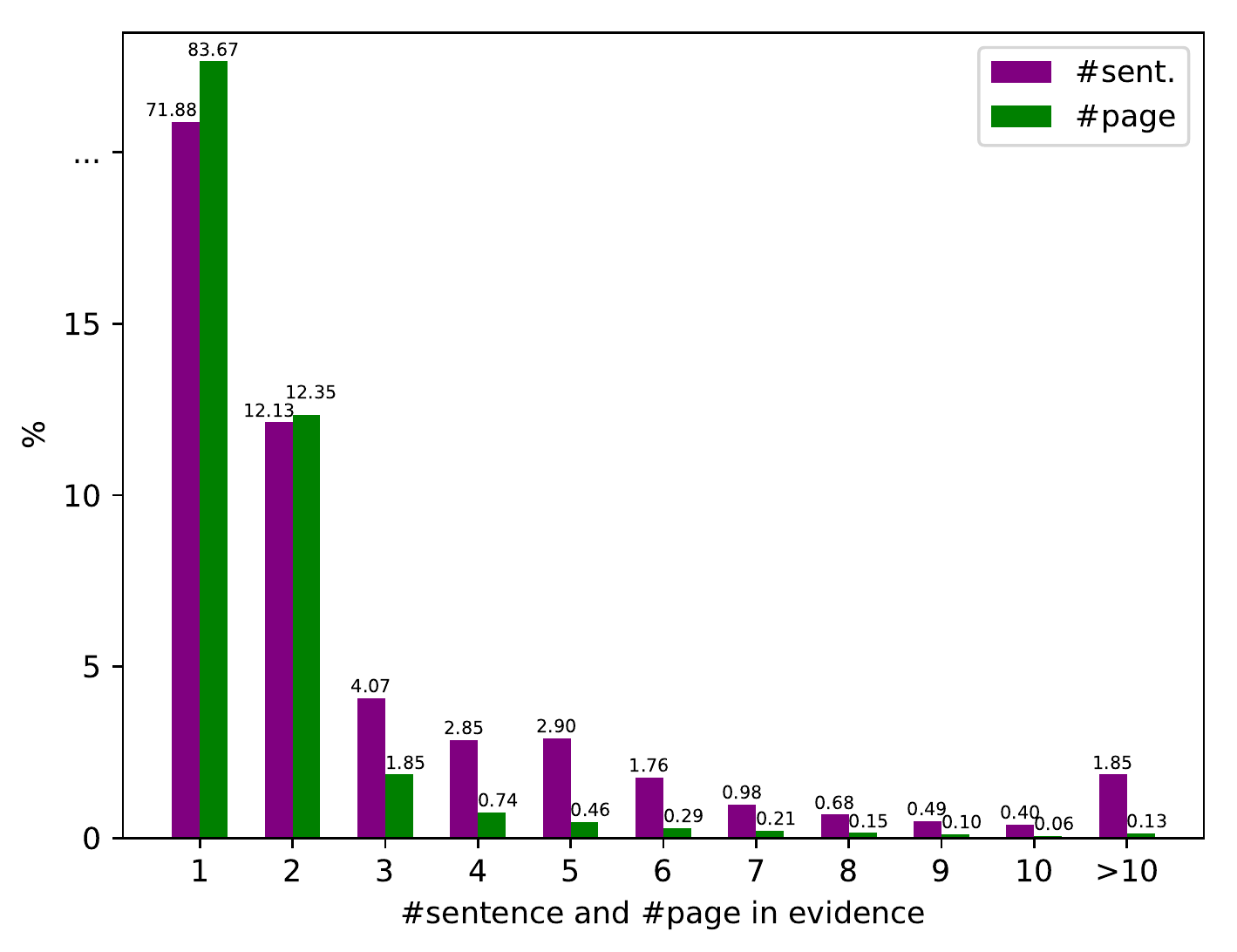}
\caption{Distribution of \#sentence and \#pages in \dataname\enspace evidence}\label{fig:evidistr}
\end{figure}

\section{Experiments}
\subsection{Setup}
\paragraph{Dataset.} In this work, we use \dataname\enspace \cite{DBLPfever05355}. The claims in \dataname\enspace were generated from  the introductory parts of about 50K Wikipedia pages of a June 2017 dump. Annotators construct claims about a single fact of the title entity with arbitrarily complex expressions and entity forms. To increase the claim complexity so that  claims would not be trivially verified, annotators adopt two routes: (i) Providing additional knowledge: Annotators can explore a dictionary of terms that were (hyper-)linked, along with their pages; (ii) Mutate claims in six ways:  negation, paraphrasing, substitution of a relation/entity with a similar/dissimilar one, and making the claims more general/specific. All resulting claims have   9.4 tokens in average. Apart from claims, \dataname\enspace also provides a Wikipedia corpus in size of about 5.4 million.

Each claim is labeled as \textsc{Supported}, \textsc{Refuted} or \textsc{NotEnoughInfo} (NEI). In addition, evidence sentences, from any wiki page, are required to be provided for   \textsc{Supported} and \textsc{Refuted}. Table \ref{tab:data} lists the data statistics. Figure \ref{fig:evidistr} shows the distributions of sentence sizes and page sizes in \dataname's evidence set. We can see that  roughly 28\% of the evidence covers more than one sentence, and approximately 16.3\% of the evidence covers more than one wiki page.

% \{1: 0.7188449107534391, 2: 0.12134771239707005, 3: 0.040741582888047946, 4: 0.028511799386074614, 5: 0.029031524581377595, 6: 0.01757320816618213, 7: 0.009809813061343814, 8: 0.006780789657468613, 9: 0.004864302999788862, 10: 0.003971025320361859, 11: 0.018523330788845396\}

This task has three evaluations: (i) \textsc{NoScoreEv} -- accuracy of claim verification, neglecting  the validity of evidence; (ii) \textsc{ScoreEv} -- accuracy of claim verification with a requirement that the predicted evidence fully covers the gold evidence for \textsc{Supported} and \textsc{Refuted};
% of providing correct evidence for \textsc{Supported} and \textsc{Refuted} (it is treated as ``correct'' once all ground truth evidence sentences are covered by the predicted evidence); 
(iii) $F_1$ -- between the predicted evidence sentences  and the ones chosen by annotators. We use the officially released evaluation scorer \footnote{https://github.com/sheffieldnlp/fever-scorer}.

\paragraph{Wiki page retrieval\footnote{Our  retrieval results are released as well.}.} For each claim, we search in the given dictionary of wiki pages in the form of \{title: sentence\_list\}, and keep the top-5 ranked pages for fair comparison with \newcite{DBLPfever05355}. Algorithm \ref{algo:docrank} briefly shows the steps of wiki page retrieval. To speed up, we first build an inverted index from words to titles, then for each claim, we only search in the titles that cover at least one claim word. 

\begin{algorithm}
\small
\SetAlgoLined
\KwIn{A claim, wiki=\{title: page\_vocab\}}
\KwOut{A ranked top-$k$ wiki titles}
 Generate entity\_mentions from the claim\;
%  Generate page.vocab() from sentence\_list\;
\While{each title}{
% read current\;
\eIf{claim.vocab$\cap$title.vocab is empty}{
discard this title
}{
title\_score = the max recall value of title.vocab in claim and in entity\_mentions of the claim\;
\eIf{title\_score = 1.0}{
title.score = title\_score
}{
page\_score = recall of claim in page\_vocab\;
title.score = title\_score + page\_score
}
}
}
Sort titles by title.score in descending order\\
\enspace
\caption{Algorithm description of wiki page retrieval for \dataname\enspace claims.}\label{algo:docrank}
\end{algorithm}
\begin{table}[t]
  \centering
%   \small
  \begin{tabular}{l|cc|cc}
  k &   \multicolumn{2}{c|}{\small\cite{DBLPfever05355} } &  \multicolumn{2}{c}{ours}\\
  & rate & acc\_ceiling & rate & acc\_ceiling\\\hline
  %0.78 0.915 0.93 0.945 0.945 0.955
1 & 25.31 & 50.21& 76.58 &84.38\\
5 & 55.30 &70.20 & 89.63 & 93.08\\
10 & 65.86 & 77.24& 91.19& 94.12\\
25 & 75.92 & 83.95&92.81&95.20\\
50 & 82.49 & 90.13&93.36&95.57\\
100 & 86.59 &91.06 &94.19&96.12\\\hline

\end{tabular}
\caption{Wikipage retrieval evaluation on dev. ``rate'': claim proportion, e.g., $x$\%, if its gold passages are fully retrieved (for ``SUPPORT'' and ``REFUTE'' only); ``acc\_ceiling'':   $\frac{x\%\cdot(\#\textsc{S}+\#\textsc{R})+\#\textsc{N}}{\#\textsc{S}+\#\textsc{R}+\#\textsc{N}}$, the upper bound of accuracy for three classes if the coverage $x$\% satisfies.}\label{tab:pageretrieval}
\end{table}

\begin{table*}
  \centering
  \begin{tabular}{lll|cc|ccc}
 & & & \multicolumn{2}{c|}{claim verification} & \multicolumn{3}{c}{evidence identification}\\
 & \multicolumn{2}{c|}{system} &  \textsc{NoScoreEv} & \textsc{ScoreEv}  & recall & precision & $F_1$\\\hline\hline
\multirow{11}{*}{\rotatebox{90}{\begin{tabular}{c}dev\end{tabular}}} &  \multicolumn{2}{l|}{MLP} & 41.86 & 19.04 & 44.22 & 10.44 & 16.89 \\
 & \multicolumn{2}{l|}{Decomp-Att} & 52.09 & 32.57 & 44.22 & 10.44 & 16.89\\\cline{2-8}
&\multirow{8}{*}{\rotatebox{90}{\begin{tabular}{c}\modelname\end{tabular}}} & coarse\&coarse & & & & &\\

&& \enspace\enspace pipeline  &35.72 & 22.26&53.75 &29.42 &33.80 \\
 & &\enspace\enspace diff-CNN & 39.22 & 21.04 & 46.88 & 43.01 & 44.86\\
&    &\enspace\enspace share-CNN & 72.32 & 50.12 & 45.55 & 40.77 & 43.03\\

\cdashline{3-8}

  %0.5364 0.7565 0.425308457711 0.458059701493 0.441076946575
 & &coarse\&fine(single) &75.65 & 52.65& 45.81&42.53 & 44.11\\
  %0.5265 0.789 0.392251243781 0.45776119403 0.422481812561
 & &coarse\&fine(two) & 78.77 & 53.64& 45.78 & 39.23 & 42.25 
\\\cdashline{3-8}
%0.499 0.6782 0.513360696517 0.524029850746 0.518640409602
%0.5343 0.7102 0.483109452736 0.527014925373 0.504108004318
&& fine\&sent-wise  &71.02 &53.43 & 52.70& \textbf{48.31}& 50.40\\
  %0.5317 0.7148 0.477288557214 0.538059701493 0.505855476519
 & &fine\&coarse &71.48 &53.17 &52.75 & 47.30& 49.87\\

  %0.5618 0.7877 0.47297761194 0.527462686567 0.498736490828
  %0.5605 0.7753 0.479139303483 0.530298507463 0.503422508546
 & &fine\&fine(two) &\textbf{78.90}& \textbf{56.16}& \textbf{53.81} &47.73 & \textbf{50.59} \\\hline\hline

\multirow{2}{*}{\rotatebox{90}{\begin{tabular}{c}test\end{tabular}}} &  \multicolumn{2}{l|}{\cite{DBLPfever05355}} & 50.91 & 31.87 & 45.89 & 10.79 & 17.47\\
& \multicolumn{2}{l|}{\modelname} &\textbf{75.99}& \textbf{54.33} & \textbf{49.91} &\textbf{44.68} & \textbf{47.15}
\end{tabular}
\caption{Performance on $dev$ and $test$ of \dataname. \modelname\enspace outperforms prior systems if vanilla CNN parameters are shared by evidence identification and claim verification subsystems. It gains more if  fine-grained representations are adopted in both subtasks.}\label{tab:results}
\end{table*}

%0.765898617512 0.896313364055 0.91198156682 0.928110599078 0.933640552995 0.941935483871

% \begin{table*}
% % \setlength{\tabcolsep}{3pt}
% %  \setlength{\belowcaptionskip}{-15pt}
% %  \setlength{\abovecaptionskip}{5pt}
%   \centering
%   \begin{tabular}{l|l|c}
%    \multicolumn{1}{c|}{claim} & \multicolumn{1}{c|}{entities} & top-5 wiki-titles\\\hline
% \multirow{5}{7cm}{The Adventures of Pluto Nash was reviewed by Ron Underwood.} & \multirow{5}{2.5cm}{Adventures; Pluto Nash; Ron Underwood } & The\_Adventures\_of\_Pluto\_Nash\\
% & & The\_Adventures\\
% & & Pluto\\
% & & The\_Adventures\_of\_Bumblefoot\\
% & & Ron\_Ron\\\hline

% \multirow{5}{7cm}{Hot Right Now is mistakenly attributed to DJ Fresh.} & \multirow{5}{2cm}{Hot; DJ Fresh} & DJ\_Fresh\\
% & & Hot\_Right\_Now\\
% & & Right\_Right\_Now\_Now\\
% & & DJ\_Fresh\_discography\\
% & & Right\_Now\\\hline
% \end{tabular}
% \caption{Examples of document retrieval result}\label{tab:docrere}
% \end{table*}

All sentences of the top-5 retrieved wiki pages are kept as evidence candidates for claims in train, dev and test. It is worth mentioning that this page retrieval step is a reasonable preprocessing which controls the complexity of  evidence searching in real-world, such as the big space -- 5.4 million -- in this work.

\paragraph{Training setup.}  All words are initialized by 300D  Word2Vec \cite{mikolov2013distributed} embeddings, and are fine-tuned during training. The whole system is trained by AdaGrad \cite{duchi2011adaptive}. Other hyperparameter values include: learning rate 0.02, hidden size 300, mini-batch size 50, filter width 3. 

\paragraph{Baselines.} In this work, we first consider the two systems reported by \newcite{DBLPfever05355}: (i) MLP: A multi-layer perceptron with one  hidden layer, based on TF-IDF cosine similarity between the claim and the evidence (all evidence sentences are concatenated as a longer text piece) \cite{DBLPRiedelASR17}; (ii) Decomp-Att
\cite{DBLParikhT0U16}: A decomposable attention model  that develops attention mechanisms to
decompose the problem into  subproblems to solve in
parallel. Note that both systems first employed an IR system to keep  top-5 relevant sentences from the retrieved top-5 wiki pages as \emph{static evidence} for claims.

We further consider the following variants of our own system \modelname:

\textbullet\enspace  \emph{Coarse-coarse}: Both evidence identification and claim verification adopt coarse-grained representations. 

To further study our system, we test this ``coarse-coarse'' in three setups: (i) ``pipeline'' -- train the two modules independently. Forward the predicted evidence to do entailment for claims; (ii) ``diff-CNN'' -- joint training with separate CNN parameters to learn sentence/claim representations; (iii) ``share-CNN'' -- joint training with shared CNN parameters.

The following variants are  in joint training.

\textbullet\enspace  \emph{Fine\&sentence-wise}:  Given the evidence with multiple sentences, a natural baseline is to do entailment reasoning for each (sentence, claim), then compose.  We do entailment reasoning between each predicted evidence sentence and the claim, generating a probability distribution over the label space $Y$. Then we sum up all the distribution vectors element-wise, as an ensemble system, to predict the label;

\textbullet\enspace  Four combinations of different grained representation learning: ``\emph{coarse\&fine(single)}'', ``\emph{coarse\&fine(two)}'', ``\emph{fine\&coarse}'' and ``\emph{fine\&fine(two)}''. ``Single'' and ``two''   refer to the single/two-channel  cases respectively.

\subsection{Results}
\paragraph{Performance of passage retrieval.}
Table \ref{tab:pageretrieval} compares our wikipage retriever with the one in \cite{DBLPfever05355}, which used a document retriever\footnote{It  compares passages and claims
as TF-IDF weighted bag-of-bigrams.} from DrQA \cite{DBLPChenFWB17}.

Our document retrieval module surpasses the competitor by a big margin in terms of the coverage of gold passages: 89.63\% vs. 55.30\% ($k=5$ in all experiments). Its powerfulness should be attributed to: (i) Entity mention detection in the claims. (ii) As wiki titles are entities, we have a bi-channel way to match the claim with the wiki page: one with the title, the other with the page body, as shown in Algorithm \ref{algo:docrank}.

% Table \ref{tab:docrere} shows two examples of document retrieval results. Even though the entity mentions can not be detected perfectly, our algorithm

\paragraph{Performance on \dataname} Table \ref{tab:results} lists the performances of baselines and the \modelname\enspace variants on \dataname\enspace (dev\&test).
From the \emph{dev} block, we observe that:

\textbullet\enspace \modelname\enspace(from ``share-CNN'') surpasses prior systems in big margins. Overall, fine-grained schemes in each subtask contribute more than the coarse-grained counterparts;

\textbullet\enspace In the three setups -- ``pipeline'', ``diff-CNN'' and ``share-CNN'' -- of \emph{coarse-coarse}, ``pipeline'' gets better scores than \cite{DBLPfever05355} in terms of evidence identification. ``Share-CNN'' has comparable $F_1$ as ``diff-CNN'' while  gaining a lot on \textsc{NoScoreEv} (72.32 vs. 39.22) and \textsc{ScoreEv} (50.12 vs. 21.04). This clearly shows that the claim verification  gains much knowledge transferred from the evidence identification module. Both ``diff-CNN'' and ``share-CNN'' perform better than ``pipeline'' (except for the slight inferiority  at \textsc{ScoreEv}: 21.04 vs. 22.26).

\textbullet\enspace  Two-channel fine-grained representations show more effective than the single-channel counterpart in claim verification (\textsc{NoScoreEv}: 78.77 vs. 75.65, \textsc{ScoreEv}: 53.64 vs. 52.65). As we expected, evidence sentences should \emph{collaborate} in inferring the truth value of the claims. Two-channel setup enables an evidence candidate aware of other candidates as well as the claim. 

\textbullet\enspace  In the last three rows of \emph{dev}, there is no clear difference among their evidence identification scores. Recall that ``sent-wise''  is essentially an ensemble system over each (sentence, claim) entailment result.  ``Coarse-grained'', instead, first sums up all sentence representation, then performs ($\sum(sentence)$, claim) reasoning. We can also treat this ``sum up'' as an ensemble. Their comparison shows that \emph{these two kinds of tricks do not make much difference}.

If we adopt ``two-channel fine-grained representation'' in claim verification, big improvements are observed in both \textsc{NoScoreEv} (+7.42\%) and \textsc{ScoreEv} (+3\%).

\begin{table*}
\setlength{\tabcolsep}{3pt}
  \centering
  \small
  \begin{tabular}{c|c|l|c|c} % index, claim, gold sents, pred sent
   \# & G/P & \multicolumn{1}{c|}{claim} & gold evidence & predicted evidence \\\hline
   \multirow{5}{*}{1} & \multirow{5}{*}{0/1} & \multirow{5}{5cm}{Telemundo is an English-language television network.} & (Telemundo, 0) & (Telemundo, 0)\\
   & & & (Telemundo, 1) & (Telemundo, 4)\\
      & & & (Telemundo, 4) & (Fourth\_television\_network, 0)\\
         & & & (Telemundo, 5) & (Fourth\_television\_network, 4)\\
            & & & (Hispanic\_and\_Latino\_Americans, 0) & \\\hline

   \multirow{4}{*}{2} & \multirow{4}{*}{1/2} & \multirow{4}{5cm}{Home for the Holidays stars a famous American actor.} & (Anne\_Bancroft, 0) & \multirow{4}{*}{$\emptyset$}\\
   & & & (Charles\_Durning, 0) & \\
      & & & (Holly\_Hunter, 0) & \\
         & & & (Home\_for\_the\_Holidays\_(1995\_film), 5) &\\\hline
         
   \multirow{3}{*}{3} & \multirow{3}{*}{0/2} & \multirow{3}{5cm}{Both hosts of Weekly Idol were born in 1983.} & (Weekly\_Idol, 0) & \multirow{3}{*}{(Weekly\_Idol, 1)}\\
   & & & (Weekly\_Idol, 1) & \\
      & & & (Defconn, 0) & \\\hline
\end{tabular}
\caption{Error cases of \modelname\enspace in \dataname. ``G/P'': gold/predicted label (``0'': refute; ``1'': support; ``2'': not enough information). To save space, we use ``(title, $i$)'' to denote the $i^{th}$ sentence in the corresponding wiki page. }\label{tab:errorexample}
\end{table*}
\begin{figure}[t]
\centering
\includegraphics[width=0.48\textwidth]{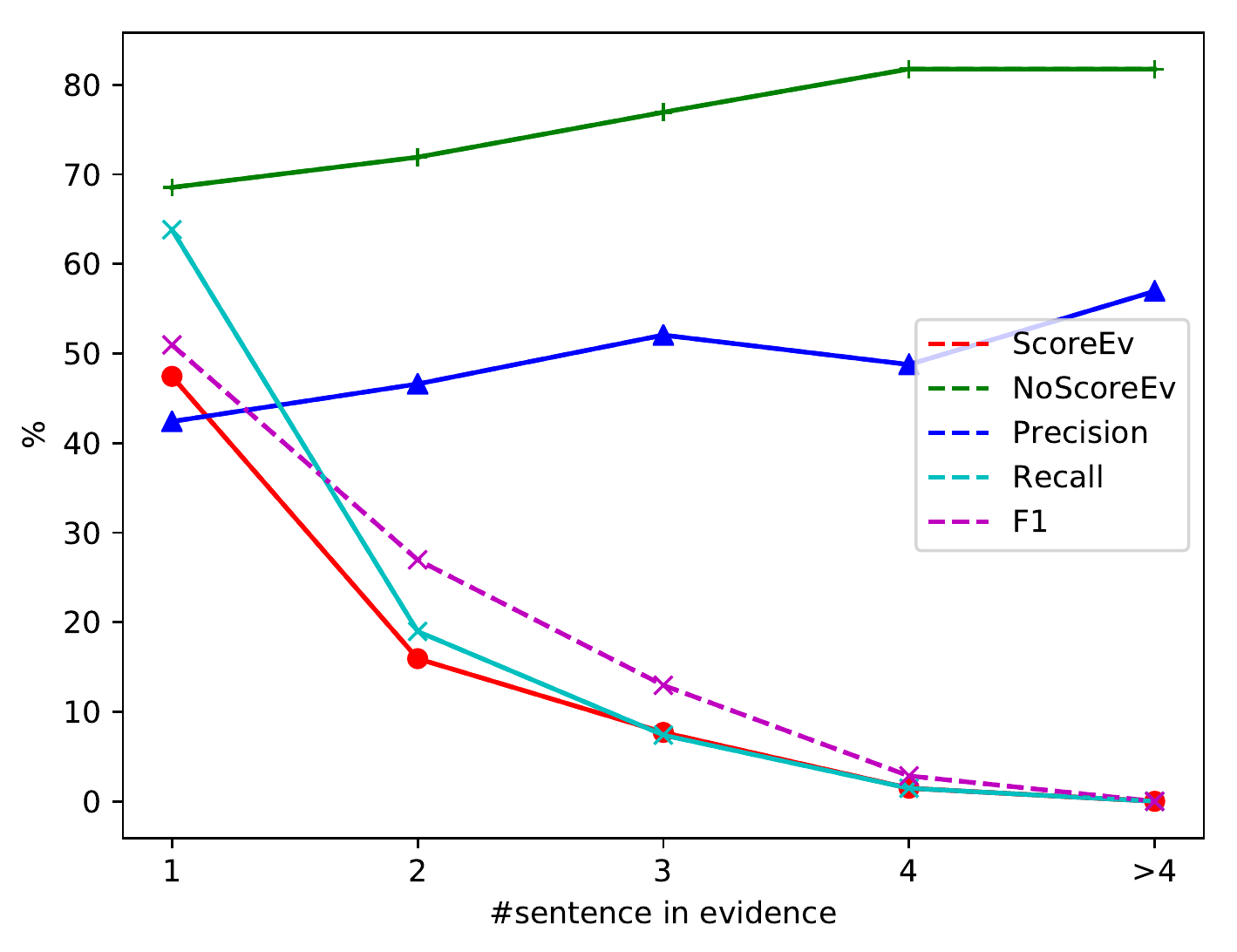}
\caption{Performance vs. \#sentence in evidence. Our system has robust precisions.  The overall performance $\textsc{NoScoreEv}$ is not influenced by the decreasing recall; this verifies the fact that the truth value of most claims can be determined by a single identified evidence sentence.}\label{fig:performdistr}
\end{figure}
In the \emph{test} block,   our system (fine\&fine(two)) beats the prior top system across all measurements by big margins -- $F_1$: 47.15 vs. 17.47;  \textsc{ScoreEv}: 54.33 vs. 31.87; \textsc{NoScoreEv}: 75.99 vs. 50.91.

In both \emph{dev} and \emph{test} blocks, we can observe that our evidence identification module consistently obtains balanced recall and precision. In contrast, the pipeline system by \newcite{DBLPfever05355} has much higher  recall than  precision (45.89 vs. 10.79). It is worth mentioning that the \textsc{ScoreEv} metric is highly influenced by the recall value, since \textsc{ScoreEv} is computed on the claim instances whose evidences are fully retrieved,  regardless of the precision. So, ideally, a system can set all sentences as evidence, so that \textsc{ScoreEv} can be promoted to be equal to \textsc{NoScoreEv}. Our system is more reliable in this perspective.

\paragraph{Performance vs. \#sent. in evidence.}

Figure \ref{fig:performdistr} shows the results of the five evaluation measures against different sizes of gold evidence sentences in test set. We observe that: (i) Our system has robust precisions across \#sentence; however, the recall decreases. This is not that surprising, since the more ground-truth sentences in evidence, the harder it is to retrieve all of them; (ii) Due to the decrease in recall, the \textsc{ScoreEv} also gets influenced for bigger \#sentence. Interestingly, high precision and worse recall in evidence with more sentences still make consistently strong  overall performance, i.e., \textsc{NoScoreEv}. This should be due to the fact that the majority (83.18\% \cite{DBLPfever05355}) of claims can be correctly entailed by a single ground truth sentence, even if any remaining ground truth sentences are unavailable.

\paragraph{Error analysis.}

The case \#1 in Table \ref{tab:errorexample} shows that our system identifies two pieces of evidence (i.e., (Telemundo, 0) and (Telemundo, 4)) correctly; however, it falsely predicts the claim label.  (Telemundo, 0): \emph{Telemundo is an American Spanish-language terrestrial television $\cdots$}. We can easily find that the keyword ``Spanish-language'' should refute the claim. However, both ``Spanish-language'' in this evidence and the ``English-language'' in the claim are unknown tokens with randomly initialized embeddings. This hints that a more careful data preprocessing may be helpful. In addition, to refute the claim, another clue comes from the combination of (Telemundo, 4) and (Hispanic\_and\_Latino\_Americans, 0). (Telemundo, 4): ``\emph{The channel $\cdots$ aimed at Hispanic and Latino American audiences}''; (Hispanic\_and\_Latino\_Americans, 0): ``\emph{Hispanic Americans and Latino Americans $\cdots$ are descendants of people from countries of Latin America and Spain.}''. Our system only retrieved (Telemundo, 4). And this clue is hard to grasp as it requires some background knowledge -- people from  Latin America and Spain usually are not treated as English-speaking.

In the case \#2, our system fails to identify any evidence. This is due to the failure of our passage retrieval module: it detects entity mentions ``Home'', ``Holidays'' and ``American'', and the top-5 retrieved passages are	``Home'',	``Home\_for\_the\_Holidays'',	``American\_Home'',	``American'' and	``Home\_for\_the\_Holidays\_(song)'', which unfortunately cover none of the four ground truth passages.
 Interestingly, (i) given the falsely retrieved passages, our system predicts ``no sentence is valid evidence'' (denoted as  $\emptyset$ in Table \ref{tab:errorexample}); (ii) given the empty evidence, our system predicts ``NoEnoughInfo'' for this claim. Both  make sense. 

In the case \#3, a successful classification of the claim requires information aggregation over the three gold evidence sentences: (Weekly\_Idol, 0): ``\emph{Weekly Idol is a South Korean variety show $\cdots$}''; (Weekly\_Idol, 1): ``\emph{The show is hosted by comedian Jeong Hyeong-don and rapper Defconn.}''; (Defconn, 0):   ``\emph{Defconn (born Yoo Dae-joon; January 6 , 1977 ) is a $\cdots$}''. To successfully retrieve  the three sentences  as a whole set of evidence is challenging in evidence identification. Additionally, this example relies on the recognition and matching of digital numbers (1983 vs. 1977), which is beyond the expressivity of word embeddings, and is expected to be handled by rules more easily.

% \paragraph{Lessons.} We summarize following lessons by this work:
% \begin{itemize}
% \item The joint training of two coupled subtasks needs to keep parameter sharing as well as parameter separation in some parts;
% \end{itemize}

% \paragraph{Training size vs. performance}

% \danchange{Add a discussion with limitations and future directions}

\section{Summary}
In this work, we build \modelname, a two-wing optimization framework to address the claim verification problem by presenting precise evidence. Differing from a pipeline system, \modelname\enspace ensures the evidence identification module and the claim verification module are  trained jointly, in an end-to-end scheme. Experiments show the superiority of \modelname\enspace in the \dataname\enspace benchmark.

\subsection*{Acknowledgments}
We thank group colleagues (Nitish Gupta and Jennifer Sheffield) and Dr. Mo Yu from IBM AI Foundations Lab for providing insightful comments and critiques. This work was supported by Contract HR0011-15-2-0025 with the US Defense Advanced Research
Projects Agency (DARPA). Approved for
Public Release, Distribution Unlimited. The views
expressed are those of the authors and do not reflect
the official policy or position of the Department of
Defense or the U.S. Government.

\bibliography{ccg,acl2018}

\begin{thebibliography}{24}
\expandafter\ifx\csname natexlab\endcsname\relax\def\natexlab#1{#1}\fi

\bibitem[{Angeli and Manning(2014)}]{DBLPngeliM14}
Gabor Angeli and Christopher~D. Manning. 2014.
\newblock Naturalli: Natural logic inference for common sense reasoning.
\newblock In \emph{Proceedings of {EMNLP}}, pages 534--545.

\bibitem[{Blacoe and Lapata(2012)}]{DBLPBlacoeL12}
William Blacoe and Mirella Lapata. 2012.
\newblock A comparison of vector-based representations for semantic
  composition.
\newblock In \emph{Proceedings of {EMNLP-CoNLL}}, pages 546--556.

\bibitem[{Bollacker et~al.(2008)Bollacker, Evans, Paritosh, Sturge, and
  Taylor}]{bollacker2008freebase}
Kurt Bollacker, Colin Evans, Praveen Paritosh, Tim Sturge, and Jamie Taylor.
  2008.
\newblock Freebase: a collaboratively created graph database for structuring
  human knowledge.
\newblock In \emph{Proceedings of SIGMOD}, pages 1247--1250.

\bibitem[{Chaganty et~al.(2017)Chaganty, Paranjape, Liang, and
  Manning}]{DBLPChagantyPLM17}
Arun~Tejasvi Chaganty, Ashwin Paranjape, Percy Liang, and Christopher~D.
  Manning. 2017.
\newblock Importance sampling for unbiased on-demand evaluation of knowledge
  base population.
\newblock In \emph{Proceedings of {EMNLP}}, pages 1038--1048.

\bibitem[{Chen et~al.(2017)Chen, Fisch, Weston, and Bordes}]{DBLPChenFWB17}
Danqi Chen, Adam Fisch, Jason Weston, and Antoine Bordes. 2017.
\newblock Reading wikipedia to answer open-domain questions.
\newblock In \emph{Proceedings of {ACL}}, pages 1870--1879.

\bibitem[{Dagan et~al.(2013)Dagan, Roth, Sammons, and Zanzoto}]{DRSZ13}
Ido Dagan, Dan Roth, Mark Sammons, and Fabio~Massimo Zanzoto. 2013.
\newblock Recognizing textual entailment: Models and applications.

\bibitem[{Duchi et~al.(2011)Duchi, Hazan, and Singer}]{duchi2011adaptive}
John Duchi, Elad Hazan, and Yoram Singer. 2011.
\newblock Adaptive subgradient methods for online learning and stochastic
  optimization.
\newblock \emph{JMLR}, 12:2121--2159.

\bibitem[{Ferreira and Vlachos(2016)}]{DBLPFerreiraV16}
William Ferreira and Andreas Vlachos. 2016.
\newblock Emergent: a novel data-set for stance classification.
\newblock In \emph{Proceedings of {NAACL}}, pages 1163--1168.

\bibitem[{Hovy et~al.(2013)Hovy, Berg{-}Kirkpatrick, Vaswani, and
  Hovy}]{DBLPHovyBVH13}
Dirk Hovy, Taylor Berg{-}Kirkpatrick, Ashish Vaswani, and Eduard~H. Hovy. 2013.
\newblock Learning whom to trust with {MACE}.
\newblock In \emph{Proceedings of {NAACL}}, pages 1120--1130.

\bibitem[{Kobayashi et~al.(2017)Kobayashi, Ishii, Hoshino, Miyashita, and
  Matsuzaki}]{DBLPKobayashiIHMM17}
Mio Kobayashi, Ai~Ishii, Chikara Hoshino, Hiroshi Miyashita, and Takuya
  Matsuzaki. 2017.
\newblock Automated historical fact-checking by passage retrieval, word
  statistics, and virtual question-answering.
\newblock In \emph{Proceedings of {IJCNLP}}, pages 967--975.

\bibitem[{LeCun et~al.(1998)LeCun, Bottou, Bengio, and
  Haffner}]{lecun1998gradient}
Yann LeCun, L{\'e}on Bottou, Yoshua Bengio, and Patrick Haffner. 1998.
\newblock Gradient-based learning applied to document recognition.
\newblock In \emph{Proceedings of the IEEE}, pages 2278--2324.

\bibitem[{Mikolov et~al.(2013)Mikolov, Sutskever, Chen, Corrado, and
  Dean}]{mikolov2013distributed}
Tomas Mikolov, Ilya Sutskever, Kai Chen, Greg~S Corrado, and Jeff Dean. 2013.
\newblock Distributed representations of words and phrases and their
  compositionality.
\newblock In \emph{Proceedings of NIPS}, pages 3111--3119.

\bibitem[{Parikh et~al.(2016)Parikh, T{\"{a}}ckstr{\"{o}}m, Das, and
  Uszkoreit}]{DBLParikhT0U16}
Ankur~P. Parikh, Oscar T{\"{a}}ckstr{\"{o}}m, Dipanjan Das, and Jakob
  Uszkoreit. 2016.
\newblock A decomposable attention model for natural language inference.
\newblock In \emph{Proceedings of {EMNLP}}, pages 2249--2255.

\bibitem[{Pasternack and Roth(2013)}]{DBLPPasternackR13}
Jeff Pasternack and Dan Roth. 2013.
\newblock Latent credibility analysis.
\newblock In \emph{Proceedings of {WWW}}, pages 1009--1020.

\bibitem[{Riedel et~al.(2017)Riedel, Augenstein, Spithourakis, and
  Riedel}]{DBLPRiedelASR17}
Benjamin Riedel, Isabelle Augenstein, Georgios~P. Spithourakis, and Sebastian
  Riedel. 2017.
\newblock A simple but tough-to-beat baseline for the fake news challenge
  stance detection task.
\newblock \emph{CoRR}, abs/1707.03264.

\bibitem[{Roth et~al.(2009)Roth, Sammons, and Vydiswaran}]{DBLPRothSV09}
Dan Roth, Mark Sammons, and V.~G.~Vinod Vydiswaran. 2009.
\newblock A framework for entailed relation recognition.
\newblock In \emph{Proceedings of {ACL}}, pages 57--60.

\bibitem[{Suchanek et~al.(2007)Suchanek, Kasneci, and
  Weikum}]{DBLPSuchanekKW07}
Fabian~M. Suchanek, Gjergji Kasneci, and Gerhard Weikum. 2007.
\newblock Yago: a core of semantic knowledge.
\newblock In \emph{Proceedings of {WWW}}, pages 697--706.

\bibitem[{Thorne et~al.(2018)Thorne, Vlachos, Christodoulopoulos, and
  Mittal}]{DBLPfever05355}
James Thorne, Andreas Vlachos, Christos Christodoulopoulos, and Arpit Mittal.
  2018.
\newblock {FEVER:} a large-scale dataset for fact extraction and verification.
\newblock In \emph{Proceedings of {NAACL}}.

\bibitem[{Vlachos and Riedel(2014)}]{DBLPVlachosR14}
Andreas Vlachos and Sebastian Riedel. 2014.
\newblock Fact checking: Task definition and dataset construction.
\newblock In \emph{Proceedings of the Workshop on Language Technologies and
  Computational Social Science@ACL}, pages 18--22.

\bibitem[{Vydiswaran et~al.(2011)Vydiswaran, Zhai, and
  Roth}]{DBLPVydiswaranZR11}
V.~G.~Vinod Vydiswaran, ChengXiang Zhai, and Dan Roth. 2011.
\newblock Content-driven trust propagation framework.
\newblock In \emph{Proceedings of {SIGKDD}}, pages 974--982.

\bibitem[{Wang et~al.(2018)Wang, Yu, Guo, Wang, Klinger, Zhang, Chang, Tesauro,
  Zhou, and Jiang}]{DBLPWangYGWKZCTZJ18}
Shuohang Wang, Mo~Yu, Xiaoxiao Guo, Zhiguo Wang, Tim Klinger, Wei Zhang, Shiyu
  Chang, Gerry Tesauro, Bowen Zhou, and Jing Jiang. 2018.
\newblock R\({}^{\mbox{3}}\): Reinforced ranker-reader for open-domain question
  answering.
\newblock In \emph{Proceedings of {AAAI}}.

\bibitem[{Wang(2017)}]{DBLPWang17}
William~Yang Wang. 2017.
\newblock ``{Liar, Liar Pants on Fire}": {A} new benchmark dataset for fake
  news detection.
\newblock In \emph{Proceedings of {ACL}}, pages 422--426.

\bibitem[{Yin et~al.(2016)Yin, Ebert, and Sch\"{u}tze}]{yinHCQA}
Wenpeng Yin, Sebastian Ebert, and Hinrich Sch\"{u}tze. 2016.
\newblock Attention-based convolutional neural network for machine
  comprehension.
\newblock In \emph{Proceedings of the {NAACL} Workshop on Human-Computer
  Question Answering}, pages 15--21.

\bibitem[{Yin and Sch{\"{u}}tze(2017)}]{DBLP00519}
Wenpeng Yin and Hinrich Sch{\"{u}}tze. 2017.
\newblock Attentive convolution.
\newblock \emph{CoRR}, abs/1710.00519.

\end{thebibliography}
\bibliographystyle{acl_natbib_nourl}
\end{document}